%% file: main.tex
\documentclass[10pt,twocolumn,letterpaper]{article}

\usepackage{comment}
\usepackage{xcolor} 
\usepackage{colortbl}
\usepackage{pifont}

\usepackage{listings}
\usepackage[accsupp]{axessibility}

\definecolor{background}{RGB}{240, 240, 240}

\lstdefinelanguage{json}{
    basicstyle=\ttfamily\footnotesize,
    backgroundcolor=\color{background},
    stringstyle=\color{blue},
    keywordstyle=\color{red},
    morestring=[b]",
    morekeywords={safe, unsafe, not_moving, stopping, wait_then_move, slowing_down, slowdown_then_speedup, speeding_up, maintaining_speed, speedup_then_slowdown},
    literate=
    	{'}{{'}}1
    	{<}{{$<$}}1
    	{>}{{$>$}}1
    	{&}{{\&}}1
    	{_}{{\_}}1
    	{^}{{\textasciicircum}}1
    	{~}{{\textasciitilde}}1
    	{=}{{=}}1
}

\usepackage{graphicx}
\usepackage{amsmath}
\usepackage{amssymb}
\usepackage{booktabs}
\usepackage{comment}
\usepackage{xcolor} 
\usepackage{colortbl}
\usepackage{pifont}
\usepackage[linesnumbered,ruled]{algorithm2e}
\usepackage{stfloats} 
\usepackage{makecell}
\usepackage{listings}

\newcommand{\cmark}{\textcolor{green}{\checkmark}}  
\newcommand{\xmark}{\textcolor{red}{\ding{55}}}  
\usepackage{wacv}              

\definecolor{wacvblue}{rgb}{0.21,0.49,0.74}
\usepackage[pagebackref,breaklinks,colorlinks,allcolors=wacvblue]{hyperref}

\def\wacvPaperID{586} 
\def\confName{WACV}
\def\confYear{2026}

\newcommand{\papername}{Instruction-Conditioned Trajectory Generation}
\newcommand{\papernameAbbrev}{iMotion-LLM}

\title{\papernameAbbrev: Instruction-Conditioned Trajectory Generation}

\author{
Abdulwahab Felemban$^{1}$ \;
Nussair Hroub$^{1}$ \;
Jian Ding$^{1}$ \;
Eslam Abdelrahman$^{1}$ 
Xiaoqian Shen$^{1}$ \\
Abduallah Mohamed$^{2}$\footnotemark[1] \quad
Mohamed Elhoseiny$^{1}$ \\
{$^{1}$King Abdullah University of Science and Technology (KAUST) \quad
$^{2}$Meta Reality Labs }\\
{\tt\small $^{1}$\{firstname.lastname\}@kaust.edu.sa}\qquad
{\tt\small $^{2}$abduallahadel@meta.com}
}

\begin{document}

\twocolumn[{%
\renewcommand\twocolumn[1][]{#1}%
\maketitle
\vspace*{-2.5em} 
\centering
\includegraphics[width=1.00\linewidth]{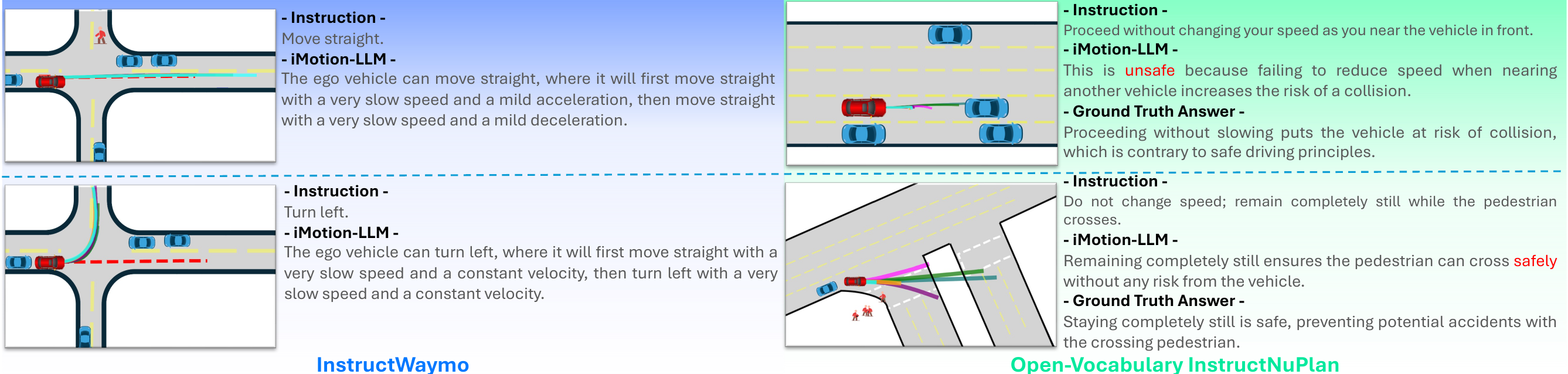}
\captionof{figure}{\papernameAbbrev{} generates feasible, safety-aligned trajectories from human instructions. It uses two datasets, InstructWaymo (direction-based) and Open-Vocabulary InstructNuPlan (safety-focused), to support instruction-conditioned generation and justification.
\vspace{1em}}

\label{teaser}
}]

{
    \renewcommand{\thefootnote}{\fnsymbol{footnote}} 
    \footnotetext[1]{This work was done outside of Meta in a personal capacity.} 
}

\input{sec/0_abstract}

\input{sec/1_intro}

\input{sec/2_related}
\input{sec/3_method}
\input{sec/4_exp}

\input{sec/5_conclusion}
{
    \small
    \bibliographystyle{ieeenat_fullname}
    \bibliography{main}
}
\clearpage

\input{sec/supp}

\end{document}

%% file: sec/0_abstract.tex
\begin{abstract}

\vspace{-0.5cm}
We introduce {\papernameAbbrev}, a large language model (LLM) integrated with trajectory prediction modules for interactive motion generation. Unlike conventional approaches, it generates feasible, safety-aligned trajectories based on textual instructions, enabling adaptable and context-aware driving behavior. It combines an encoder-decoder multimodal trajectory prediction model with a pre-trained LLM fine-tuned using LoRA, projecting scene features into the LLM input space and mapping special tokens to a trajectory decoder for text-based interaction and interpretable driving. To support this framework, we introduce two datasets: 1) InstructWaymo, an extension of the Waymo Open Motion Dataset with direction-based motion instructions, and 2) Open-Vocabulary InstructNuPlan, which features safety-aligned instruction-caption pairs and corresponding safe trajectory scenarios. Our experiments validate that instruction conditioning enables trajectory generation that follows the intended condition. {\papernameAbbrev} demonstrates strong contextual comprehension, achieving 84\% average accuracy in direction feasibility detection and 96\% average accuracy in safety evaluation of open-vocabulary instructions. This work lays the foundation for text-guided motion generation in autonomous driving, supporting simulated data generation, model interpretability, and robust safety alignment testing for trajectory generation models. Our code, pre-trained model, and datasets are available at: \href{https://vision-cair.github.io/iMotion-LLM/}{vision-cair.github.io/iMotion-LLM/}.

\vspace{-0.6cm}
\end{abstract}

%% file: sec/1_intro.tex
\section{Introduction}
\label{sec:intro}

Trajectory prediction is a core component of autonomous driving pipelines, enabling the ego vehicle to anticipate surrounding agents and plan safe maneuvers \citep{hu2023_uniad}. Large-scale benchmarks such as the Waymo Open Motion Dataset (WOMD) \citep{ettinger2021large} and the Waymo Sim Agents Challenge \citep{NEURIPS2023_b96ce67b} have advanced prediction accuracy and realism, but they provide limited controllability: models forecast futures from past motion without explicit mechanisms to direct agent behavior. As autonomous systems mature, the ability to \emph{control} generated trajectories, particularly through natural language, becomes essential for scenario generation, safety testing, and interpretability.

Recent research has explored language-conditioned traffic and trajectory generation, including LANG-TRAJ~\cite{LANGTRAJ}, InteractTraj~\cite{InteractTraj}, ProSim-Instruct~\cite{promsim_2024}, and others~\cite{lctgen,ctg,ctg++,feng2023trafficgen}. While these methods demonstrate progress in scene-level realism or simulation-based evaluation, none provide a rigorous framework for measuring controllability under diverse language inputs. In contrast, our work establishes \emph{Instruction-Conditioned Trajectory Generation} as a task: we pair large-scale real-world datasets with textual instructions, introduce a direct metric for instruction adherence (IFR), and compare against strong trajectory prediction backbones to ensure preserving trajectory quality.

Our framework lays the foundation for next-generation AV systems that can reason and act safely based on language. 
While NuPlan~\cite{nuplan} enables sampling of challenging cases through scenario mining, our setting emphasizes semantically controlled trajectories. This is similar in spirit to concurrent efforts~\cite{promsim_2024, LANGTRAJ}, but distinguishes itself by providing a rigorous evaluation of controllability. As a result, developers can not only represent challenging scenarios, but also manipulate agent motion to stress-test downstream planning under diverse conditions, where prior work such as LCTGen~\cite{lctgen} has shown that such manipulation can significantly degrade the performance of policy-based planners.

\noindent\textbf{Our contributions are:}
\begin{itemize}
    \item We introduce two new datasets: \textbf{InstructWaymo} and \textbf{Open-Vocabulary InstructNuPlan}, which augment open-loop driving scenarios with direction-based and free-form, safety-aligned instructions.
    \item We develop instruction-conditioned models, including an LLM-based approach that integrates feasibility and safety reasoning for trajectory generation.
    \item We propose \textbf{Instruction Following Recall (IFR)}, an effective metric to evaluate adherence to instructions.
\end{itemize}

%% file: sec/2_related.tex
\section{Related Work}

\noindent{\textbf{Multimodal Large Language Models.}} Large Language Models (LLMs) have significantly advanced in recent years~\citep{GPT2,bert,GPT3,llama2,touvron2023llama,GPT-4}, with models like GPT-4~\citep{GPT-4} demonstrating remarkable abilities in generating coherent, contextually relevant text across numerous domains. With the strong performance of LLMs, there is an emergence of multi-modal LLMs (MLLMs)~\citep{flamingo}, which extend the LLMs with reasoning abilities across diverse modalities. Notable works include VisualGPT~\cite{chen2022visualgpt},  Flamingo~\citep{flamingo},  MiniGPT-4~\citep{chen2023minigpt,zhu2023minigpt}, LLaVA~\citep{llava,llava1.5}, and InstructBLIP~\citep{instructblip}. These works used visual instruction tuning to align with human intentions. There are some extensions that focus on detection and segmentation~\citep{zhu2023minigpt,wang2024visionllm,lai2023lisa,bai2023qwen}, videos~\citep{li2023videochat,video-llama,video-chatgpt}, and 3D~\citep{3D-LLM,xu2023pointllm,guo2023point}. Our work focuses on MLLMs for motion prediction tasks.


\noindent{\textbf{Trajectory Prediction Models for Driving Scenarios.}} Trajectory prediction analyzes agents' historical tracks on maps to forecast joint future positions. Early approaches used LSTMs~\citep{alahi2016social,lstm} for encoding agent states and CNNs~\citep{cui2019multimodal,gilles2021home,salzmann2020trajectron++} for rasterized scene images. GNNs~\citep{chen2022scept,huang2022multi,mo2022multi} later improved agent interaction modeling. Transformer-based models like SceneTransformer~\citep{ngiam2021scene} and WayFormer~\citep{wayformer} enhanced prediction efficiency but focused mainly on encoding vectorized representations. Motion Transformer~\citep{shi2022motion,shi2024mtr++} and GameFormer~\citep{huang2023gameformer} achieved better accuracy through improved decoding, while MotionLM~\citep{motionllm} adopted LLM-like structures without incorporating language reasoning capabilities. Earlier work also examined social-awareness in~\citep{saadatnejad2022sattack}.


\noindent{\textbf{Multimodal Large Language Models for Autonomous Driving.}} The emergence of LLMs has sparked adaptations for autonomous driving~\citep{chen2023driving,dewangan2023talk2bev,hu2023gaia,huang2022language}. GPT-Driver~\citep{mao2023gpt} and SurrealDriver~\citep{jin2023surrealdriver} demonstrate LLMs' transformative impact on motion planning and maneuver generation. However, most methods focus on text or image inputs, overlooking vector representation's benefits for motion prediction. Vector representation abstractly captures essential driving scenario information. Like Driving with LLMs~\citep{chen2023driving}, we integrate LLMs with vector-based data, but while~\citep{chen2023driving} introduced QA-focused benchmarks representing motion as single quantized actions (acceleration, braking, steering), we focus on multi-modal multi-agent trajectories, better aligning with existing trajectory prediction modules for safer, more reliable motion prediction.


\noindent{\textbf{Conditional and Promptable Trajectory Generation.}}  
Several works explore controllable or instruction-driven trajectory generation. \textbf{InteractTraj}~\cite{InteractTraj} and \textbf{TrafficGen}~\cite{feng2023trafficgen} generate synthetic data with controllability over traffic density or interactions but evaluate mainly on realism. Diffusion-based methods like CTG and CTG++~\cite{ctg,ctg++} add goal- or LLM-guided constraints, while LCTGen~\cite{lctgen} edits scenarios from free-form text, though without reproducible benchmarks or controllability metrics. Closer to our setting, LANGTRAJ~\cite{LANGTRAJ} and ProSim-Instruct~\cite{promsim_2024} pair trajectories with instructions, but support limited instruction types and report only realism-based simulation metrics. In contrast, we provide large-scale datasets, a dedicated controllability metric (IFR), and rigorous benchmarking against strong trajectory prediction backbones, while also enabling textual justification of feasibility and safety alignment with driving scenarios, establishing Instruction-Conditioned Trajectory Generation as a standalone task.

%% file: sec/3_method.tex
\section{Motion Instruction Datasets}
\label{dataset}
\subsection{InstructWaymo Dataset}
\label{sec_InstructWaymo}

\begin{table}[t!]
  \centering
  \caption{Direction categories with their corresponding presence proportion in the train set.}
  \label{tab:direction_stat}
   \vspace{-3.5mm}
   \scalebox{0.70}
  {
  \begin{tabular}[t]{ccccc}
    \toprule
    \textbf{Stationary} & \textbf{Straight}  & \textbf{Right} & \textbf{Left} & \textbf{Left U-turn} \\
    \midrule
    16,748 & 863,910 & 184,286 & 221,762 & 4,575 \\
    \bottomrule
  \end{tabular}
  }
  \vspace{-6mm}
\end{table}

InstructWaymo offers a new perspective on WOMD \cite{WaymoOpenDataset} by making trajectory generation instructable and language descriptive. Inspired by the mAP calculation used in WOMD, which evaluates model performance across various driving behaviors, we designed a module that categorizes future motion based on direction, speed, and acceleration. 
InstructWaymo uses future direction information as instructions alongside detailed motion descriptions such as two-step direction, speed, and acceleration, which are presented as captions to be used as output text for driving behavior interpretation by models. Additionally, the feasibility of each instruction (direction) is calculated, adding an extra layer of comprehension by identifying feasible and infeasible directions for each driving scenario.

InstructWaymo is publicly available as a script to augment WOMD. This data augmentation is applied to approximately 400K driving scenarios, which are preprocessed similarly to the GameFormer preprocessing method. Each scenario includes up to 32 neighboring agents, with a total of 33 agents, including the ego agent. Each agent in the scene is considered a focal agent (the ego-view agent), resulting in 4.2M samples across different focal agents. Among these, 1.3M samples involve a unique vehicle as the focal agent with a valid trajectory and detected instructions.

\noindent \textbf{Direction.}
Direction is fundamental for instructing navigation; we adopted the WOMD direction bucketing script to obtain five conceivable direction conditions encompassing five classes listed in Table \ref{tab:direction_stat} with their statistics. The statistics shows a bias toward some behaviors like moving straight. We use driving directions as instructions in the experiments on InstructWaymo. 


\noindent \textbf{Speed and Acceleration.}
Following the intuition used in \citep{mohamed2022social}, we categorize trajectories of moving vehicles based on speeds and relative change in speeds. For that, we defined 5-speed categories and 9-acceleration categories; the suggested upper threshold and the categories are listed in the supplementary.


\begin{figure*}[t!]
\centering
\vspace{-3mm}
\includegraphics[width=0.95\linewidth]{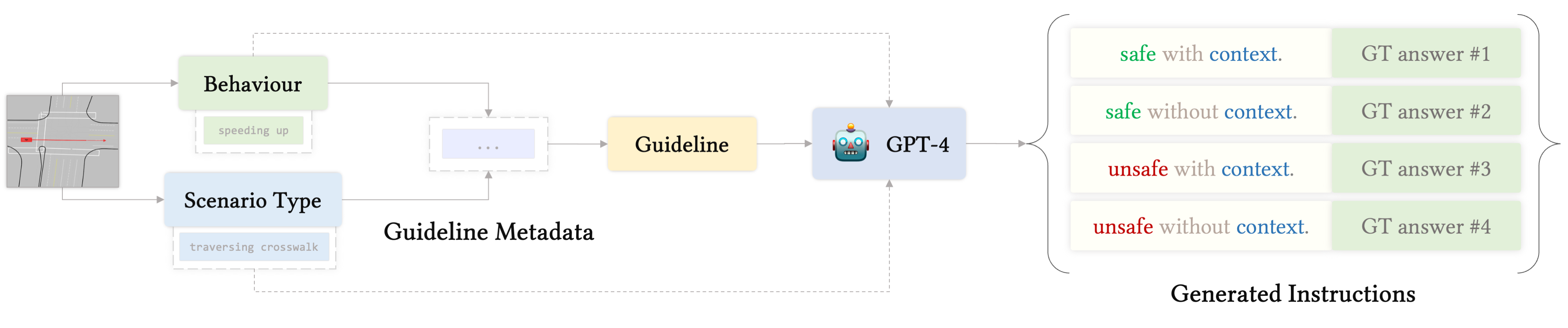}
\vspace{-2mm}
\caption{Open-Vocabulary InstructNuPlan pipeline for generating safe/unsafe instruction-caption pairs using GPT-4o mini, guided by scenario type, motion behavior, and safety metadata.}

\vspace{-4mm}
\label{fig:nuplan_preprocess}
\end{figure*}

\noindent \textbf{Feasibility of directions.}
We define the feasibility of directions into three categories: 1) Ground-Truth direction (GT), which is based on the ground-truth future trajectory and hence is always assumed to be a feasible direction; 2) Feasible directions (F), which are derivable directions but not the GT direction; 3) Infeasible directions (IF), which is the complement set of feasible directions. To assess feasibility, we consider a set of candidate destination lanes relative to the ego vehicle's current location and heading. These candidate destinations are possible locations on associated lanes within a range determined by the vehicle's speed (maximum range $r$). This range is calculated based on a maximum speed increase of 15 km/h within 8 seconds, not exceeding the road speed limit, and within a maximum range of 60 meters, indicating a safe range of acceleration or deceleration. For the feasibility of staying stationary, we detect if the vehicle current speed is within 65 km/h, allowing it to slow down to stop in 8 seconds. 
These heuristics are grounded in general safe driving practices and are configurable in the codebase that is publicly available.

\noindent \textbf{LLM Instruction and caption.}
Using the extracted attributes, we construct a templated pair comprising an input instruction and an output caption for the LLM. The input instruction specifies the target (final) direction the ego vehicle should reach. The output caption—generated autoregressively—covers the target direction and an interpretable execution plan: two intermediate wayfinding steps, along with indicative speeds and accelerations that realize the final direction. This pairing makes the model’s adherence to the instruction explicit and verifiable.

\subsection{Open-Vocabulary InstructNuPlan Dataset}
\label{sec_InstructNuplan}

Open-Vocabulary InstructNuPlan extends NuPlan \cite{nuplan} to explore instruction-conditioned trajectory generation beyond predefined instruction sets. Unlike InstructWaymo, which focuses on structured direction-based commands, this dataset introduces diverse open-vocabulary instructions, enabling more flexible and context-aware motion generation. It also emphasizes safety alignment, distinguishing between safe and unsafe behaviors to evaluate an autonomous vehicle’s ability to follow safe instructions while rejecting unsafe ones. This dataset serves as a benchmark for assessing instruction-following capabilities in real-world driving scenarios. The full process of generating open-vocabulary instruction-caption pairs for a driving scenario is illustrated in Figure~\ref{fig:nuplan_preprocess}. The scenario type is provided in the original NuPlan dataset \cite{nuplan}.

\noindent \textbf{Selected Scenario Types.}  
To capture diverse driving contexts, we select 14 scenario types from the NuPlan dataset, spanning a wide range of real-world decision-making challenges such as interactions with pedestrians, vehicles, and road infrastructure. The chosen scenarios include: Accelerating at crosswalk, Traversing crosswalk, Waiting for pedestrian to cross, Following lane (with/without/with slow lead), Stopping with lead, Starting protected/non-protected turns (cross/non-cross), Behind bike, Behind long vehicle, and Traversing intersection. We sample an equal number of examples per type, yielding 7,119 training and 124 testing scenarios. From each, we generate multiple safe and unsafe instruction–caption pairs using GPT-4o mini, resulting in 115,097 training and 2,078 testing pairs. Following WOMD conventions, we use 1.1 seconds of past motion to predict 8 seconds into the future.

\noindent \textbf{Safety Grounding and Instruction Generation.}  
To generate safety-aligned instruction–caption pairs, we construct metadata that explicitly defines safe and unsafe behaviors for each scenario type, following established traffic rules and autonomous driving best practices. Each scenario is annotated with up to 10 safe and 10 unsafe behaviors. Given a scenario, we load its metadata, determine the ego vehicle’s future motion, and match it against predefined behavior categories (\textit{not moving, stopping, waiting then moving, slowing down, speeding up, slowing down then speeding up, speeding up then slowing down, maintaining speed}). From this, we retrieve guidelines specifying safe and unsafe outcomes. For instance, in a \textit{waiting for a pedestrian to cross} scenario, if the ego vehicle’s behavior is \textit{not moving}, the safe instruction is “Do not move; the vehicle should remain stationary while a pedestrian is crossing”, whereas any movement is considered unsafe. This process ensures that instructions generated by GPT-4o mini are both context-aware and safety-aligned.

\noindent \textbf{Context-Enriched Instruction Generation.}  
Beyond labeling safe vs. unsafe behaviors, we generate instructions in two formats to test the model’s ability to handle explicit and implicit cues:
1.	\textbf{With context:} explicitly references scene elements (e.g., “Slow down before the intersection due to oncoming traffic.”).
2.	\textbf{Without context:} abstract, context-agnostic phrasing (e.g., “Reduce speed to 30 km/h.”).
This dual formulation enables a more comprehensive assessment of contextual understanding. For captions, we always include contextual details to remain aligned with the instruction.

\section{{\papernameAbbrev}}
\label{sec:method}

\begin{figure*}[t!]
\centering
\includegraphics[width=0.9\linewidth]{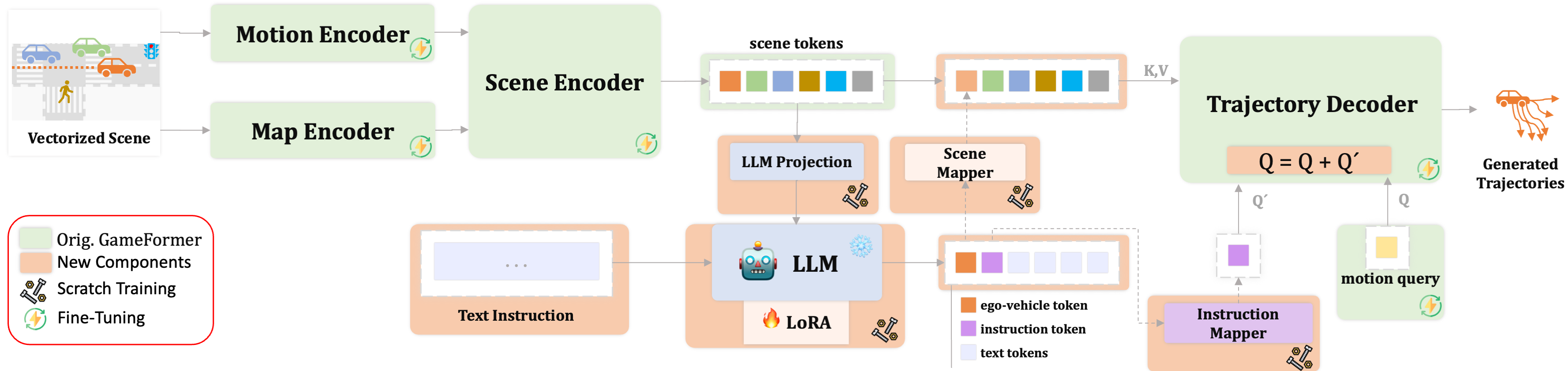}
\caption{
Given a textual instruction and scene context embeddings, {\papernameAbbrev} employs an LLM Projection module to map the encoded scene token embeddings from the Scene Encoder into the LLM input space. The LLM then generates an updated ego-vehicle token, an instruction token, and caption text tokens. The instruction token is projected into a query $Q'$, while the ego-vehicle token is projected to form the key and value representing the ego-vehicle embedding, which is subsequently used by the Multimodal Trajectory Decoder. Newly added components are highlighted in orange.
}
\label{main_figure}
\end{figure*}

\subsection{Baselines}
\label{sec_revisting_existing_models}

\noindent\textbf{Non-Conditional Baselines.}
Transformer-based trajectory prediction models like GameFormer~\citep{gameformer} and MTR~\citep{mtr} typically consist of two components: a Scene Encoder and a Multimodal Trajectory Decoder. The Scene Encoder fuses observed map and agent information into scene context embeddings (\textit{scene tokens}) $ \in \mathbb{R}^{R \times d_{\text{scene}}}$, where $d_{\text{scene}}$ is the embedding dimension. Each embedding maintains a fixed correspondence to its source (e.g., ego vehicle motion is always the first token), which preserves spatial grounding. The decoder then uses cross-attention with these embeddings as keys and values ($K,V$), and $M$ learnable queries $Q \in \mathbb{R}^{M \times d_{\text{scene}}}$ to predict a Gaussian Mixture Model (GMM) of future multimodal trajectories. These two modules—Scene Encoder and Multimodal Trajectory Decoder—form the backbone of such models and are illustrated in Figure~\ref{main_figure} in green.

\noindent{\textbf{Conditional Baselines.}} \label{subsec:conditional_decoder}
cGAN \citep{mirza2014conditional} introduces a foundational approach for incorporating conditional labels in image generation. Inspired by this, we enable decoding in originally non-conditional transformer-based baselines to be conditioned on a given instruction label. In the decoder, we introduce an additional learnable query, $Q' \in \mathbb{R}^{1 \times d_{\text{scene}}}$, which is fused with the motion generation queries, $Q$. The fusion is performed by adding $Q'$ element-wise to each of the $M$ motion queries. When training the conditional baselines without the use of LLM components (i.e., the orange components in Figure~\ref{main_figure}), $Q'$ is learned through a simple embedding layer that takes a categorical class as input. When integrating an LLM with the base model—leveraging its capability to process diverse natural language instructions—$Q'$ is instead derived from the LLM's output embeddings. Further details on training the conditional baseline are provided in the pseudo-code in the supplementary material.

\subsection{Text-Guided Trajectory Generation}
To enable natural-form text-based guidance and interaction, we incorporate an LLM into pre-trained conditional baselines consisting of a Scene Encoder and a Multimodal Trajectory Decoder.
To enable this integration design, illustrated in Figure \ref{main_figure}, four main blocks are required:
1.LLM Projection module.
2.LLM itself.
3.Scene Mapper.
4.Instruction Mapper.

\noindent \textbf{LLM Projection.}
Inspired by Vision Language Models (VLMs)~\citep{dai2023instructblip,zhu2024minigpt}, we employ a simple linear projection layer to map input scene embeddings to $\mathbb{R}^{R\times d_{LLM}}$, aligning with the LLM embedding dimension $d_{LLM}$.

\noindent \textbf{LLM.}
All projected scene embeddings, along with the input text instruction, are fed into the LLM, which generates output tokens representing the ego-vehicle token, the instruction token, and additional textual tokens that form an output caption after grounding the input instruction.

\noindent \textbf{Scene Mapper.}
We map the instruction-grounded ego token from $\mathbb{R}^{1 \times d_{\text{LLM}}}$ back to $\mathbb{R}^{1 \times d_{\text{scene}}}$ using a multilayer perceptron (MLP), referred to as the Scene Mapper. This mapped token replaces the corresponding ego token and is combined with the remaining keys and values from the Scene Encoder, which bypass the LLM. Together, they serve as the keys and values for the Multimodal Trajectory Decoder.

\noindent \textbf{Instruction Mapper.}
Following the Scene Mapper, we project the instruction token back to the motion generation model’s embedding space ($\mathbb{R}^{1 \times d_{\text{scene}}}$) using an MLP. The resulting vector is then fused with the motion query $Q$ via element-wise addition

\noindent \textbf{Output Caption.}
Along with generating scene and instruction tokens, the LLM has the capability of outputting text that describes how the instruction is executed, a textual decision ('[Accept]' or '[Reject]') to indicate whether an instruction is feasible or not, and a safety justification for accepting or rejecting the instruction.

%% file: sec/4_exp.tex
\section{Experiments}

\subsection{Datasets Setup }
\noindent \textbf{Waymo Open Motion Dataset (WOMD).} For training 
For training unconditional baseline models, we use the WOMD dataset \cite{WaymoOpenDataset}, which consists of more than 400K driving scenarios. Considering each agent out of 32 agents as the focal agent, we construct 4.2M training examples to reproduce the results of GameFormer \cite{gameformer} and MTR \cite{mtr}. We consider a trajectory window of 9 seconds, with 1.1 seconds of observed history.


\noindent \textbf{InstructWaymo.}
Using InstructWaymo, we train conditional models on 1.3M samples, each pairing the ego vehicle with one of five direction instructions. Each scenario is annotated with a ground-truth (GT) instruction, additional feasible (F) alternatives, and infeasible (IF) instructions.
For non-LLM conditional baselines, we train from scratch using GT only. In contrast, {\papernameAbbrev} additionally incorporates IF instructions during fine-tuning to strengthen feasibility awareness.
For evaluation, we use 1,500 examples balanced across GT/F/IF. Trajectory training follows WOMD settings, except we restrict the dataset to cases where the ego agent is a vehicle.

\noindent \textbf{Open-Vocabulary InstructNuPlan.}
For open-vocabulary fine-tuning and testing of {\papernameAbbrev}, we use this dataset with 115,097 training examples and 2,078 test examples. We split the test samples into four categories: safe without context, safe with context, unsafe without context, and unsafe with context. We use the same trajectory training setup as WOMD.

\noindent \textbf{Metrics.}  
Traditional trajectory metrics such as minADE/minFDE capture geometric accuracy but do not assess whether a model follows the instructed direction. We therefore introduce \textit{Instruction Following Recall (IFR)}, which measures the fraction of generated trajectories whose final direction matches the instructed direction. We extract directions from predictions using the same direction-extraction module as for ground truth.
Formally, for $N$ scenarios and $M$ generated trajectories per scenario,
\begin{equation}
\mathrm{IFR}
= \frac{1}{N}\sum_{i=1}^{N}\;\frac{1}{M}\sum_{j=1}^{M}
\mathbb{1}\!\left[\,D^{\text{pred}}_{i,j} = D^{\text{instr}}_{i}\,\right],
\end{equation}
where $D^{\text{instr}}_{i}$ is the instructed direction bucket for scenario $i$, and $D^{\text{pred}}_{i,j}$ is the direction bucket extracted from the $j$-th generated trajectory.
Higher IFR indicates stronger instruction adherence, while minADE/minFDE quantify trajectory quality. In addition, we evaluate feasibility and safety detection with classification accuracy, and report justification-quality metrics in the supplementary material.


\begin{figure*}[!t]
    \centering
    \includegraphics[width=0.95\linewidth]{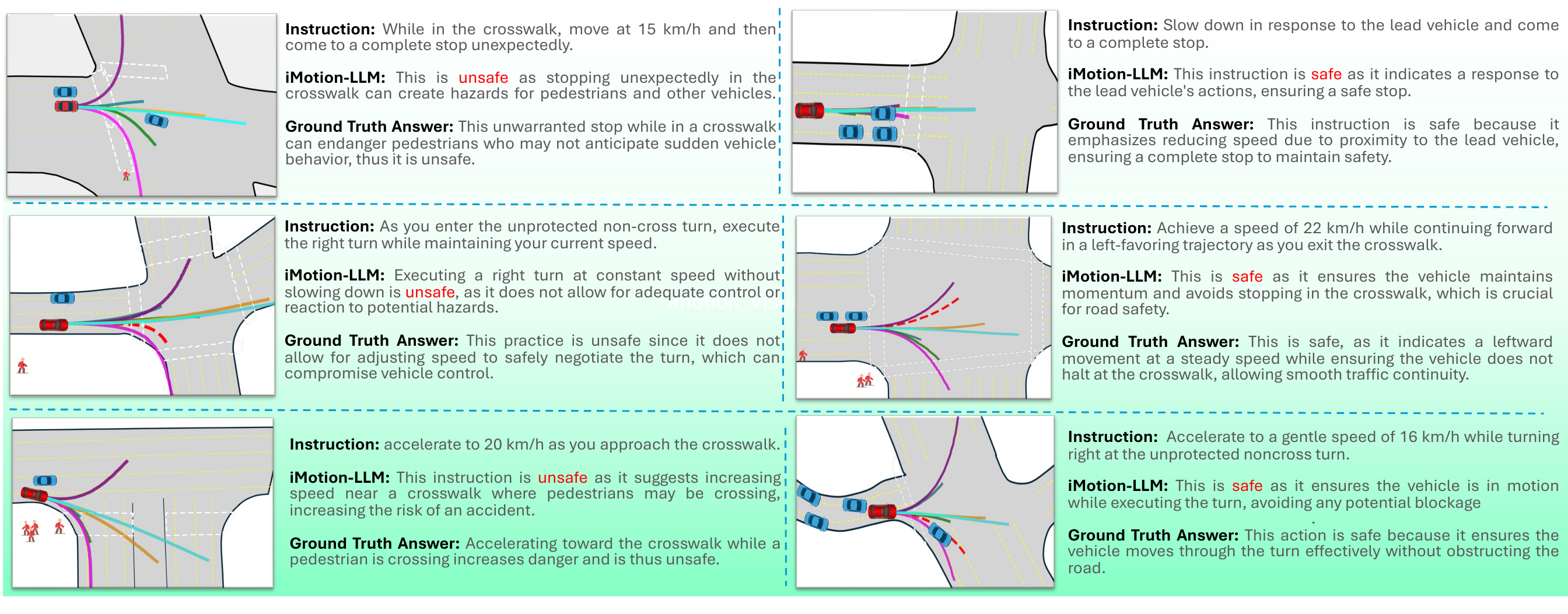}
    \caption{\textbf{\papernameAbbrev{} Qualitative Results.} Qualitative results showcasing three unsafe instructions (left) and three safe instructions (right). The results demonstrate \papernameAbbrev{}'s capability to generate relevant trajectories, assess the safety of given instructions, and provide reasoning for its decisions.}
    \label{teaser_more_nuplan_examples}
\end{figure*}
\subsection{Qualitative Results of \papernameAbbrev{}}
In Figure ~\ref{teaser_more_nuplan_examples}, \papernameAbbrev{} generates context-aware trajectories for \textbf{unsafe} (left) and \textbf{safe} (right) instructions, effectively assessing safety and providing grounded, diverse predictions with safety-aligned justifications.


\noindent{\textbf{Unsafe Instructions Results.}}  
The first unsafe case (top-left) instructs an abrupt stop in a crosswalk, risking pedestrians. \papernameAbbrev{} flags this, with five of six trajectories avoiding the stop. The second case (middle-left) instructs maintaining speed while turning, risking loss of control. \papernameAbbrev{} detects this, generating varied-speed trajectories and reinforcing the need to decelerate. The third case (bottom-left) instructs acceleration near a crosswalk, ignoring pedestrians. \papernameAbbrev{} flags the risk, producing both cautious and aggressive trajectories, with textual reasoning aiding safety filtering.

\noindent{\textbf{Safe Instructions Results.}}  
The first safe case (top-right) instructs slowing for a slow-moving lead vehicle and stopping. \papernameAbbrev{} correctly follows this, generating adaptive trajectories and reinforcing its understanding of dynamic agent interactions. 
The second safe case (middle-right) instructs a left-favoring trajectory at a controlled speed. \papernameAbbrev{} correctly follows this, recognizing the unprotected turn even though it is not explicitly mentioned in the ground truth response, showcasing strong contextual reasoning. The final safe case (bottom-right) instructs accelerating to 16 km/h while turning. \papernameAbbrev{} classifies this as safe, generating diverse paths. Notably, one trajectory takes a left turn, highlighting the model’s ability to explore multiple reasonable options.

\subsection{Results on InstructWaymo}

\begin{table}[t!]
  \centering
  \caption{\textbf{InstructWaymo Results.} Instruction Following Recall (IFR) on feasible instruction types: Ground-Truth (GT), Feasible (F), minADE and minFDE, and the feasibility detection capability of different models.}
  \label{tab:ifr_results}
  \footnotesize
  \setlength{\tabcolsep}{4pt}
  \scalebox{0.89}
  {
  \begin{tabular}{lcccc}
    \toprule
    \textbf{Model} & \textbf{GT-IFR} $\uparrow$ & \textbf{F-IFR} $\uparrow$ & \textbf{minADE/minFDE} ↓ & \makecell{\textbf{Feasibility}\\\textbf{Detection}} \\
    \midrule
    GameFormer \cite{gameformer}                    & 66.89  & 15.21       & 0.78 / 1.64 & \xmark \\
    MTR \cite{mtr}                          & 53.46    &  16.64    & 0.74 / 1.62 & \xmark \\
    \midrule
    C-GameFormer              & 83.10      &  48.70  & \textbf{0.65} / \textbf{1.20} & \xmark \\
    C-MTR                     & 64.29      &  52.22  & 0.67 / 1.39 & \xmark \\
    ProSim-Instruct & 86.32 & 24.91 & 3.52 / 3.9 & \xmark \\
    {\papernameAbbrev}     & \textbf{87.30} & \textbf{52.24} & 0.67 / 1.25 & \cmark \\
    \bottomrule
  \end{tabular}
  }
\end{table}

\noindent \textbf{Task and Metrics.}
To evaluate \papernameAbbrev{}, we assess the ability of trajectory generation models to follow a given \textit{categorical direction} using the Instruction-Following Recall (IFR) metric. Additionally, we evaluate the model's ability to reason about instruction feasibility by measuring the accuracy of accepting feasible instructions (GT or F) and rejecting infeasible ones (IF). To complement the instruction-following evaluation, we also report minADE and minFDE on GT-instruction examples to assess the quality of the generated trajectories. These conventional metrics ensure that the model not only follows instructions but also produces accurate and realistic motion plans.


\begin{table}[t!]
\vspace{-2mm}
  \centering
  \caption{\textbf{Open-Vocabulary InstructNuPlan Results.} comparison of two different types of input complex instructions in NuPlan: Safe and Unsafe. We also split the evaluation into instructions with and without context. IFR is not reported for unsafe instructions, as corresponding ground-truth trajectories are unavailable.}
  \vspace{-2mm}
  \label{tab:feasibility_results_nuplan}
   \scalebox{0.80}
  {
  \begin{tabular}[t]{lccc}
    \toprule
    \textbf{I/P Instruction} & \textbf{Context}  & \textbf{Accuracy (\%)} & \textbf{IFR} $\uparrow$ 
    \\\midrule
    Safe & \xmark & 96.61  & 68.96 \\
    Safe  & \cmark & \textbf{98.39}  & \textbf{70.22}  \\
    Unsafe & \xmark & 95.00  & N/A \\
    Unsafe  & \cmark & 96.88  & N/A \\
    \bottomrule
  \end{tabular}
  }
  \vspace{-2mm}
\end{table}

\noindent \textbf{Evaluated Models.}
Although unconditional models such as GameFormer \cite{gameformer} and MTR \cite{mtr} do not take instructions or direction categories as input, we evaluate them as lower-bound baselines to assess the effectiveness of the conditional baselines and \papernameAbbrev{}. This comparison is not intended to demonstrate superiority over unconditional models, as the tasks differ fundamentally, but rather to analyze whether incorporating instruction-following capabilities enables instructable trajectory generation, as implemented in C-GameFormer, C-MTR, and \papernameAbbrev{}.

For training \papernameAbbrev{}, we use a conditionally adapted and pretrained version of GameFormer, referred to as C-GameFormer. We adopt a mixed training strategy: 70\% of the data consists of ground-truth (GT) instructions paired with their corresponding trajectories, while the remaining 30\% consists of infeasible (IF) instructions, for which no ground-truth trajectory is provided—only the textual instruction indicating that it should be rejected. During training, the text loss is always backpropagated, while the trajectory loss is applied only when a corresponding ground-truth trajectory is available (i.e., for GT instructions). Table~\ref{tab:ifr_results} summarizes the evaluation results of the models.


\noindent \textbf{Results of Non-Conditional Baselines.}
While non-conditional MTR and GameFormer can generate realistic driving trajectories, their generated trajectories do not necessarily follow the ground truth direction (GT-IFR), as shown in Table \ref{tab:ifr_results}. Moreover, as expected, the recall of trajectories generated in other feasible directions (F-IFR) is very low, as these models were neither optimized to cover a diverse range of directions for the same driving scenario nor prompted to do so during inference.

\begin{figure}[t!]
  \centering
  \includegraphics[width=0.5\linewidth]{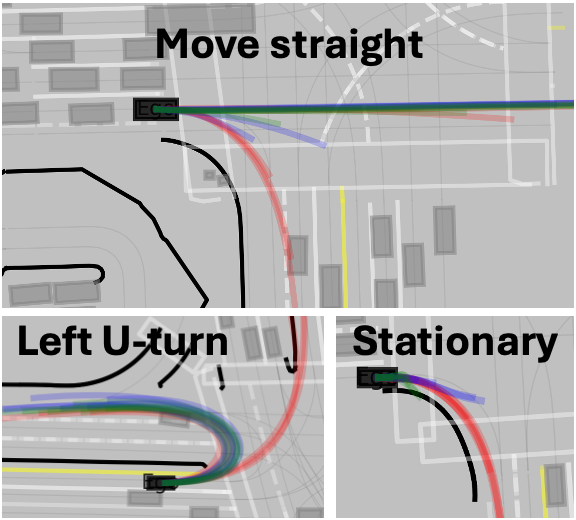}
  \caption{
  Qualitative comparison of three challenging InstructWaymo scenarios. Non-conditional GameFormer in
  \textcolor{red}{red}, C-GameFormer in \textcolor{blue}{blue}, and {\papernameAbbrev} in \textcolor{green}{green}.
  }
  \label{fig:qual}
  \vspace{-1.5em}
\end{figure}

\noindent \textbf{Results of Conditional Baselines.}  
Table~\ref{tab:ifr_results} shows that making models conditional, namely C-GameFormer and C-MTR, improve IFR for both GT and F instructions, while also reducing trajectory error as indicated by lower minADE and minFDE scores. This demonstrates that providing the intended direction as an input signal helps models generate more accurate and instruction-aligned trajectories. For comparison with recent approaches, we evaluate ProSim-Instruct, pretrained on ProSim-Instruct-450k, which is based on WOMD and uses instructions similar to InstructWaymo; in our setting, however, we evaluate single-agent conditioning under the same scenarios and instructions. Although ProSim-Instruct achieves comparable GT-IFR, it suffers from degraded trajectory quality as measured by minADE/minFDE, reflecting its emphasis on standard realism-oriented metrics. In contrast, \papernameAbbrev{} maintains strong trajectory quality while further improving IFR, and introduces a key advantage: the ability to generate textual justifications and detect instruction feasibility. Qualitative results in Figure~\ref{fig:qual} illustrate these improvements: GameFormer fails to cover diverse directions due to its non-conditional nature, C-GameFormer enables instruction adherence, and \papernameAbbrev{} achieves the best alignment, including in challenging cases such as \textit{stop}. Notably, \papernameAbbrev{} reaches 52.24\% F-IFR despite not observing such instructions during training, compared to only 24.91\% for ProSim-Instruct, highlighting both the gains of our approach and the persistent challenge of generalizing to alternative controllable driving modes in instruction-conditioned trajectory generation.

\subsection{Results on Open-Vocabulary InstructNuPlan}
\noindent \textbf{Task and Metrics.}
In open-vocabulary instructions, \papernameAbbrev{} is the only model that can address this challenge through its language processing capabilities. We evaluate its performance by assessing its ability to determine instruction safety in a given driving scenario, the primary metric for this task. Additionally, we compute Instruction-Following Recall (IFR) based on the alignment of safe instructions with the ground-truth trajectory, as they are derived from it and the scenario type.


\noindent \textbf{Safety Detection.}
As shown in Table \ref{tab:feasibility_results_nuplan}, the model demonstrates strong performance in distinguishing between safe and unsafe instructions, with slightly higher accuracy in detecting unsafe instructions. This may be attributed to certain examples containing explicitly unsafe directives, such as ``speed up without stopping for the pedestrian crossing." Interestingly, the model maintains high safety detection accuracy even without additional context. When context is included, accuracy improves slightly, which aligns with the evaluation setup where contextual information serves as an additional hint rather than an inherent advantage.

\noindent \textbf{Open-Vocabulary Instruction Following.}
The IFR performance in Table \ref{tab:feasibility_results_nuplan} indicates that \papernameAbbrev{} retains a reasonable level of IFR, surpassing the non-conditional GameFormer baseline reported on InstructWaymo (Table \ref{tab:ifr_results}), though not reaching the performance of \papernameAbbrev{} on InstructWaymo. Given that the instructions are in natural language and do not necessarily correspond to a predefined direction category, unlike in InstructWaymo where IFR is computed based on structured directional commands, the results remain reasonable. Additionally, qualitative examples suggest that the model exhibits greater diversity in the generated trajectories on Open-Vocabulary InstructNuPlan compared to InstructWaymo.

\subsection{Ablation Studies}
We conducted a series of ablation studies to analyze the influence of different key components on \papernameAbbrev{} performance.

\noindent\textbf{Impact of Class-Balanced Sampling.}
Given the strong class imbalance in motion directions, we compare results with and without class-balanced sampling for GT instructions for fine-tuning \papernameAbbrev{}. As shown in Table~\ref{tab:class_balance}, balanced sampling leads to consistent improvements in both GT-IFR and F-IFR across most instruction classes.

\begin{table*}[h!]
  \centering
  \caption{Comparison of LLMs on InstructWaymo (left) and Open-Vocabulary InstructNuPlan (right). GT/F/IF-Acc denotes the feasibility detection accuracy. Safe/Safe+Context refer to instructions without and with context, and Avg Acc reports overall safety detection accuracy.}

  \label{tab:llm_combined}
  \footnotesize
  \setlength{\tabcolsep}{4pt}
  \vspace{-1.2em}
  \resizebox{0.8\linewidth}{!}{%
  \begin{tabular}{lccccccccc}
    \toprule
     &\multicolumn{6}{c}{\textbf{InstructWaymo}} & \multicolumn{3}{c}{\textbf{Open-Vocabulary InstructNuPlan}} \\
    \textbf{LLM} & \textbf{minADE/FDE} & \textbf{GT-IFR} & \textbf{GT-Acc} & \textbf{F-IFR} & \textbf{F-Acc} & \textbf{IF-Acc} 
     & \textbf{Safe IFR} & \textbf{Safe+context IFR} & \textbf{Avg Acc} \\
    \cmidrule(r){1-1}
    \cmidrule(r){2-7}
    \cmidrule(r){8-10}
    LLaMA-7B~\cite{llama2}        & \textbf{0.67}/\textbf{1.25} & \textbf{87.30} & 97.33 & \textbf{52.24} & 62.27 & 92.73
                     & \textbf{68.96} & \textbf{70.22} & 96.25 \\
    
    Mistral-7B~\cite{jiang2023mistral7b}    & \textbf{0.67}/\textbf{1.25} & 86.61 & \textbf{97.93} & 47.23 & \textbf{64.87} & \textbf{95.00}
                    & 63.69 & 63.95 & \textbf{97.25} \\

    LLaMA-1B~\cite{grattafiori2024llama3herdmodels}      & 0.69/1.33 & 79.31 & 97.20 & 35.36 & 63.93 & 91.13
                      & 62.05 & 63.20 & 96.50 \\

    Vicuna-7B~\cite{vicuna2023}    & 0.70/1.35 & 80.81 & 94.47 & 36.56 & 39.80 & 06.13
                     & 62.26 & 62.75 & 96.25 \\
    \bottomrule
  \end{tabular}
  }
\end{table*}

\begin{table}[h!]
  \centering
    \caption{\textbf{Effect of class-balanced sampling.} GT-IFR and F-IFR across instruction classes, comparing fine-tuning \papernameAbbrev{} with and without class-balanced sampling.}
  \label{tab:class_balance}
  \footnotesize
  \setlength{\tabcolsep}{4pt}
  \scalebox{0.90}
  {
  \begin{tabular}{lcccc}
    \toprule
    \textbf{Class} 
    & \multicolumn{2}{c}{\textbf{Without Balance}} 
    & \multicolumn{2}{c}{\textbf{With Balance}} \\
    & \textbf{GT-IFR} $\uparrow$ & \textbf{F-IFR} $\uparrow$ 
    & \textbf{GT-IFR} $\uparrow$ & \textbf{F-IFR} $\uparrow$ \\
    \cmidrule(r){1-1}
    \cmidrule(r){2-3}
    \cmidrule(r){4-5}
    Stationary   &  27.61 & 0.83    & 51.00 & 16.89   \\
    Straight     &  92.11 & 26.61   & \textbf{98.83} & \textbf{70.39}  \\
    Left         &  80.78 & 19.56   & 95.94 & 68.11 \\
    Right        &  82.28 & 19.83   & 97.61 & 68.56  \\
    Left U-turn  &  \textbf{94.00} & \textbf{32.33}   & 93.11 & 37.28  \\
    \bottomrule
  \end{tabular}
  }
  \vspace{-1.0em}
\end{table}

\paragraph{LLM Backbone Comparison.}
To assess the impact of the LLM's capabilities, we evaluate \papernameAbbrev{} with various backbones. While LLaMA-7B achieves the best instruction-following recall overall, Mistral-7B shows stronger safety justification accuracy in NuPlan. Table \ref{tab:llm_combined} presents a full comparison. All other experiments use LLaMA-7B as the LLM model.

\paragraph{Effect of LLM Mapper Depth.}
We compare projection modules of varying depth. As shown in Table \ref{tab:proj_ablation}, the combination of linear and 2-layer MLP achieves the best performance across all metrics except GT-Acc and F-Acc.

\paragraph{Training Strategy for Backbone Modules.}
We evaluate different fine-tuning configurations: freezing the backbone, fine-tuning only the trajectory decoder, only the scene encoder, and full fine-tuning. Training only the decoder provides the best balance between performance and stability, as shown in Table \ref{tab:freeze_ablation}.

%% file: sec/5_conclusion.tex
\section{Limitations}

\noindent\textbf{IFR Metric.}  
Our IFR metric uses discrete directional thresholds (as in WOMD), which capture basic direction-following but not alignment under complex or open-vocabulary instructions. VLM-based evaluation is a promising alternative, though current VLMs struggle with BEV-rendered inputs.

\noindent\textbf{Dataset Bias.}  
The datasets may inherit biases from preprocessing and class imbalance. Vectorized maps introduce lossy simplifications, and Open-Vocabulary InstructNuPlan relies on guided prompts to reduce hallucination, which may limit generalization to unconstrained instructions.  

\noindent\textbf{Multi-Agent Support.}  
{\papernameAbbrev} is designed for single-agent controllability and shows reduced performance in multi-agent prediction. Richer datasets and models explicitly supporting interactive instructions are needed; initial multi-agent results are included in the supplementary.  

\begin{table}[ht!]
  \centering
  \caption{\textbf{Projection Layer Ablation.} Comparison of different input/output projection architectures. 2L and 4L denote 2-layer and 4-layer MLPs.}
  \label{tab:proj_ablation}
  \footnotesize
  \vspace{-0.5em}
  \setlength{\tabcolsep}{4pt}
  \resizebox{\linewidth}{!}{%
  \begin{tabular}{cccccccc}
    \toprule
    \textbf{In-proj} & \textbf{Out-proj} & \textbf{minADE/FDE ↓} 
    & \textbf{GT-IFR ↑} & \textbf{GT-Acc ↑} 
    & \textbf{F-IFR ↑} & \textbf{F-Acc ↑} 
    & \textbf{IF-Acc ↑} \\
    \cmidrule(r){1-2}
    \cmidrule(r){3-8}
    Linear     & Linear     & 0.78 / 1.58 & 80.78 & \textbf{97.40} & 36.83 & \textbf{65.87} & 92.13 \\
    Linear     & MLP 2L     & \textbf{0.67 / 1.25} & \textbf{87.30} & 97.33 & \textbf{52.24} & 62.27 & \textbf{92.73} \\
    MLP 2L     & MLP 2L     & 0.82 / 1.77 & 66.49 & 97.13 & 21.42 & 63.20 & 92.13 \\
    MLP 4L     & MLP 4L     & 0.76 / 1.57 & 68.73 & 97.13 & 16.49 & 63.93 & 91.07 \\
    \bottomrule
  \end{tabular}
  }
  \vspace{-1em}
\end{table}

\begin{table}[ht!]
  \centering
  \caption{\textbf{Ablation on Backbone Fine-Tuning.} Comparison of different fine-tuning strategies in \papernameAbbrev{}.}
  \label{tab:freeze_ablation}
  \footnotesize
  \setlength{\tabcolsep}{4pt}
  \resizebox{\linewidth}{!}{%
  \begin{tabular}{lcccccc}
    \toprule
    \textbf{Training Strategy} 
    & \textbf{minADE/FDE ↓} 
    & \textbf{GT-IFR ↑} & \textbf{GT-Acc ↑} 
    & \textbf{F-IFR ↑} & \textbf{F-Acc ↑} 
    & \textbf{IF-Acc ↑} \\
    \cmidrule(r){1-1}
    \cmidrule(r){2-7}
    Frozen                 & 0.78 / 1.64 & 68.46 & 97.00 & 27.19 & 63.00 & 92.33 \\
    Traj Decoder Only      & \textbf{0.67 / 1.25} & \textbf{87.30} & \textbf{97.33} & \textbf{52.24} & 62.27 & \textbf{92.73} \\
    Scene Encoder Only     & 0.70 / 1.33 & 82.32 & \textbf{97.33} & 39.04 & 64.53 & 90.33 \\
    Fully Finetuned        & 0.76 / 1.57 & 74.53 & 97.20 & 41.70 & \textbf{66.07} & 91.60 \\
    \bottomrule
  \end{tabular}
  }
  \vspace{-1em}
\end{table}

\noindent\textbf{Generalization.}  
The lower performance on alternative feasible instructions suggests limited robustness. Strong results on ground-truth instructions may partly reflect biases from future log conditioning rather than true controllable generation.  

\noindent\textbf{Computation.}  
LLM-based models incur higher inference costs than lightweight predictors. While advances in sparse attention~\cite{zhu2024star}, quantization~\cite{kim2023fast}, activation sparsity~\cite{liu2024qsparse}, and hardware-aware frameworks~\cite{wu2024flightllm} improve efficiency, LLM-based modeling is currently more suited to offline applications.

\section{Conclusion}

In conclusion, we present {\papernameAbbrev}, a large language model for Instruction-Conditioned Trajectory Generation. By leveraging textual instructions, it produces contextually relevant, safety-aligned trajectories while interpreting diverse commands. Through LoRA fine-tuning of a pretrained LLM, {\papernameAbbrev} maps scene features into the LLM input space, enabling grounded and interpretable motion reasoning. Despite the added compute cost of LLMs, the improvements in interpretability, flexibility, and safety alignment justify the trade-off for safety-critical offline applications such as simulation. Our results show that {\papernameAbbrev}, guided by InstructWaymo, effectively aligns with feasible instructions and rejects infeasible ones. Open-Vocabulary InstructNuPlan further enhances safety-aware decision-making. This work lays the foundation for text-guided motion generation in autonomous driving, enabling simulation of challenging scenarios and robust safety reasoning. Further dataset and training details are provided in the supplementary, and all code and models are publicly released for reproducibility.

\newpage
\section{Acknowledgments}
Special thanks to Habib Slim, Lucas Bezerra, and Yahia Battach for their valuable assistance during the preparation of this work. For computer time, this research used Ibex managed by the Supercomputing Core Laboratory at King Abdullah University of Science \& Technology (KAUST) in Thuwal, Saudi Arabia.

%% file: sec/supp.tex
\renewcommand{\thefigure}{S\arabic{figure}}
\setcounter{figure}{0}
\renewcommand{\thesection}{S\arabic{section}}
\setcounter{section}{0}
\renewcommand{\thetable}{S\arabic{table}}
\setcounter{table}{0}




\section{More Details on Data Preparation}
\subsection{Speed and Acceleration Categories}
The set consists of five different speed categories ranging from very slow to very fast, and a set of acceleration or deceleration levels ranging from mild to extreme, including no acceleration (i.e., constant velocity). These thresholds were designed heuristically but can be easily adapted to match real-life practical applications. For this study, however, they are sufficient to demonstrate the model's bias and comprehension regarding different speed categorizations. Tables \ref{tab:speed} and \ref{tab:accelerate} show the thresholds used.

\begin{table}[hbtp]
  \centering
  \caption{Speed categories and upper thresholds.}
  \label{tab:speed}
  \scalebox{0.65}{
    \begin{tabular}{@{}lccccc}
      \toprule
      \textbf{Speed category} & \textbf{Very slow} & \textbf{Slow} & \textbf{Moderate} & \textbf{Fast} & \textbf{Very fast} \\
      \cmidrule(r){1-1}
        \cmidrule(r){2-6}
      Threshold (km/h) & 20 & 40 & 90 & 120 & $>$ 120 \\
      \bottomrule
    \end{tabular}
  }
\end{table}

\begin{table}[hbtp]
  \centering
  \caption{Acceleration/deceleration categories and thresholds.}
  \label{tab:accelerate}
  \scalebox{0.55}{
    \begin{tabular}{@{}lccccc}
      \toprule
      \textbf{Accel./Decel. category} & \textbf{Constant velocity} & \textbf{Mild} & \textbf{Moderate} & \textbf{Aggressive} & \textbf{Extreme} \\
      \cmidrule(r){1-1}
        \cmidrule(r){2-6}
      \textbf{Threshold (km/h increase in 8s)} & 6 & 25 & 46 & 65 & $>$ 65 \\
      \bottomrule
    \end{tabular}
  }
\end{table}

\subsection{Calculation of the Directions}
\label{sec:calc_direction}
\begin{figure}[h!]
\centering
\includegraphics[width=1.0\linewidth]{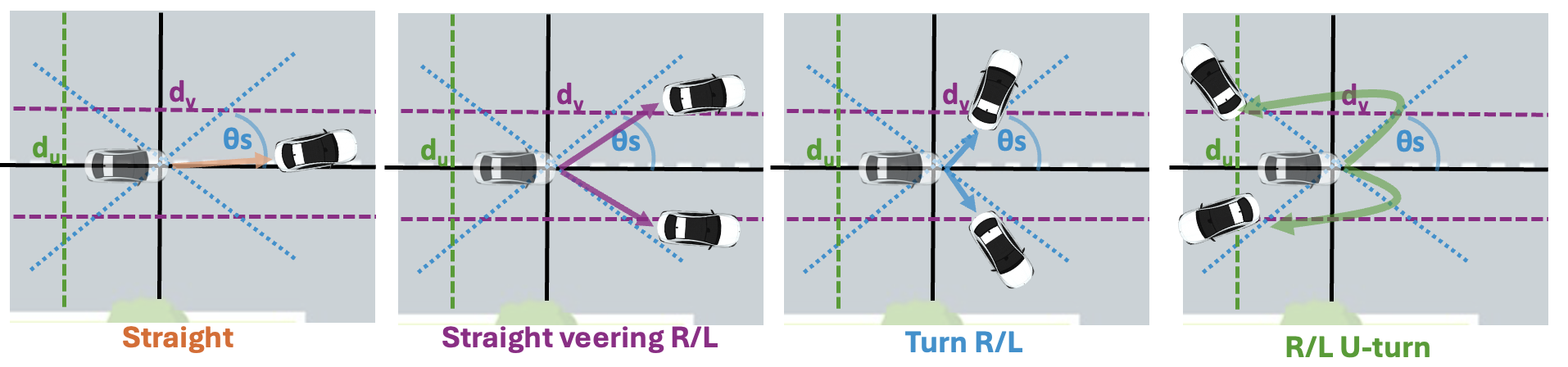}
\caption{
Illustrative examples of directions categories.}
\vspace{-4mm}
\label{directions_figure}
\end{figure}

Following the illustration shown in Fig. \ref{directions_figure}, motion direction is measured based on the relative heading angle between a time step and a future target step. We calculate direction solely based on trajectory information; the heading angle is calculated using two consecutive trajectory discrete samples. If the maximum future speed is within a threshold of $\text{v}_\text{stationary}=2 \text{m/s}$, and the vehicle traveled a distance within $\text{d}_\text{stationary} = 5 \text{m}$, the vehicle is considered stationary. Otherwise, the vehicle is moving straight if the relative heading is within $\theta_\text{s}=30$  degrees. But if the longitudinal displacement is greater than $\text{d}_\text{v} = 5 \text{m}$, it is categorized as straight veering right/left. If the relative heading exceeds $\theta_\text{s}$, and the latitudinal shift is less than $\text{d}_\text{u}=5 \text{m}$ in the opposite direction, it is considered as turning right/left. Otherwise, it is a U-turn. Right and left directions are distinguished based on the sign of the relative heading. Fig. \ref{directions_figure} illustrates the different classes. This definition is based on WOMD definition used during evaluation.

\subsection{Data Heuristics Sensitivity}
InstructWaymo was constructed by merging fragmented lane segments and applying heuristics to generate ground-truth (GT) and feasible/infeasible (F/IF) instruction buckets. GT buckets follow \href{https://github.com/waymo-research/waymo-open-dataset/issues/755}{Waymo’s official logic}, which has a verified accuracy of 98\%, while F/IF buckets rely on tunable thresholds—achieving 91\% accuracy with the default configuration. Sensitivity analysis shows that look-ahead distance has the largest effect (84\% at 100\,m vs. 79\% at 500\,m), whereas speed increase and stop heuristics have a minor impact (ranging from 89\% to 92\%). All parameters are configurable to reflect varying driving styles. For the test split, we manually verified all samples to ensure 100\% accuracy. InstructNuPlan is less sensitive to heuristic variation, as it is grounded in scenario types rather than directional feasibility. Reported numbers are based on 20 evaluated samples per category for each setup, except for InstructWaymo test data, which was fully verified and filtered.

\subsection{Evaluation of Open-Vocabulary InstructNuPlan Instructions-Reasoning Data}
To construct Open-Vocabulary InstructNuPlan, we prompted GPT with meta-actions and contextual safety indicators, specifying what is safe or unsafe to do in each driving situation. For this, we discretized speed, direction, and acceleration into buckets, and then manually assigned high-level labels indicating safety or risk. This iterative design was intended to reduce GPT hallucinations and improve consistency.

\begin{figure*}[h!]
\centering
\includegraphics[width=1.0\linewidth]{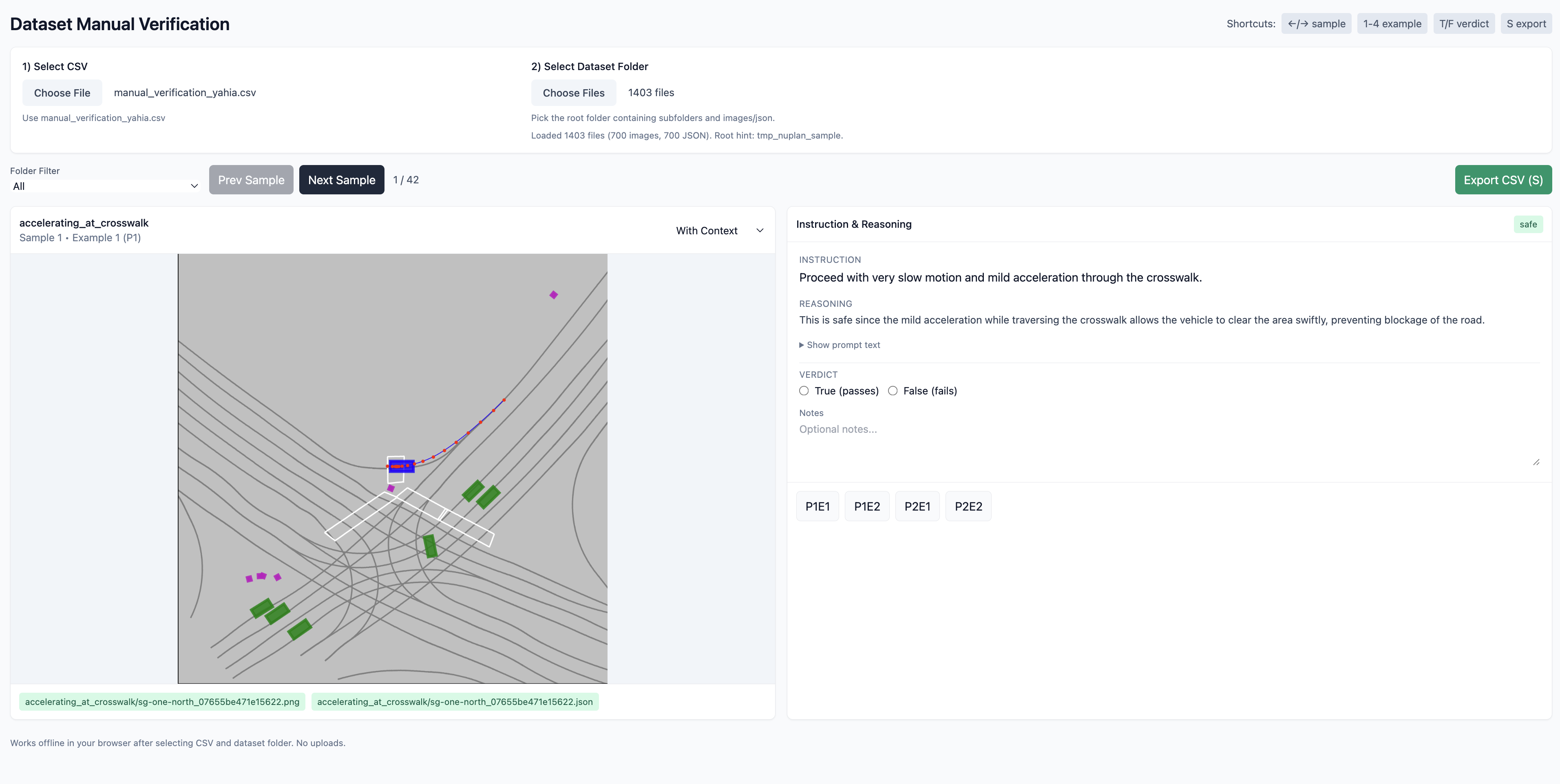}
\caption{Example of the verification web application used to evaluate Open-Vocabulary InstructNuPlan dataset.}
\vspace{-4mm}
\label{manual_verif_figure}
\end{figure*}

To verify the data quality, we developed a lightweight web-based tool that allows human annotators to inspect samples and check whether the generated instruction and reasoning align with the driving scenario. Figure~\ref{manual_verif_figure} illustrates this tool. Following the same evaluation framework used in prior work, we validated a 10\% random sample of ground-truth instructions and justifications using both human annotators and GPT-5. The results show an average score of \textbf{9.4} (GPT) and \textbf{9.0} (human), with a Pearson correlation of \textbf{0.88} between them. This high agreement confirms that the generated justifications are strongly aligned with the trajectories and provide reliable supervision for training and evaluation. 

Overall, Open-Vocabulary InstructNuPlan enables both promptable trajectory evaluation and assessment of LLM-generated justifications, bridging instruction-following with trajectory quality.

\begin{figure*}[h!]
\centering
\includegraphics[width=1.0\linewidth]{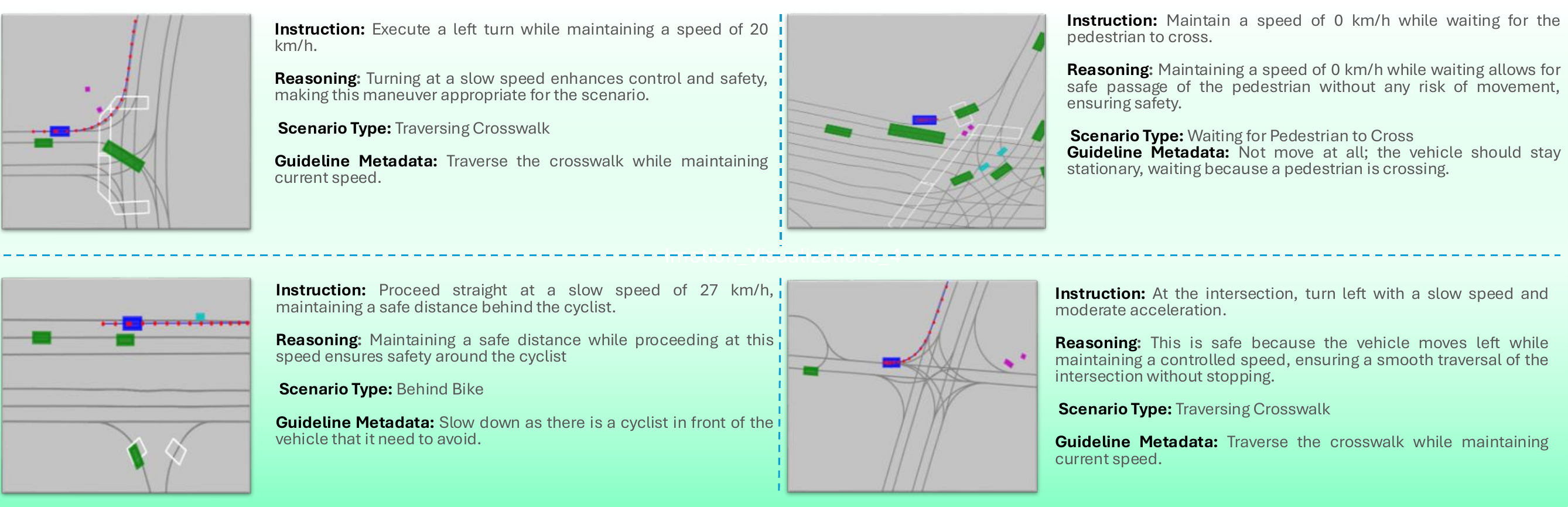}
\caption{Sampled examples from the dataset showing the instruction used to prompt the model, the LLM's reasoning about the scenario, scenario type, and the guideline metadata.}
\label{data_quality}
\end{figure*}

\begin{figure*}[h!]
\centering
\includegraphics[width=1.0\linewidth]{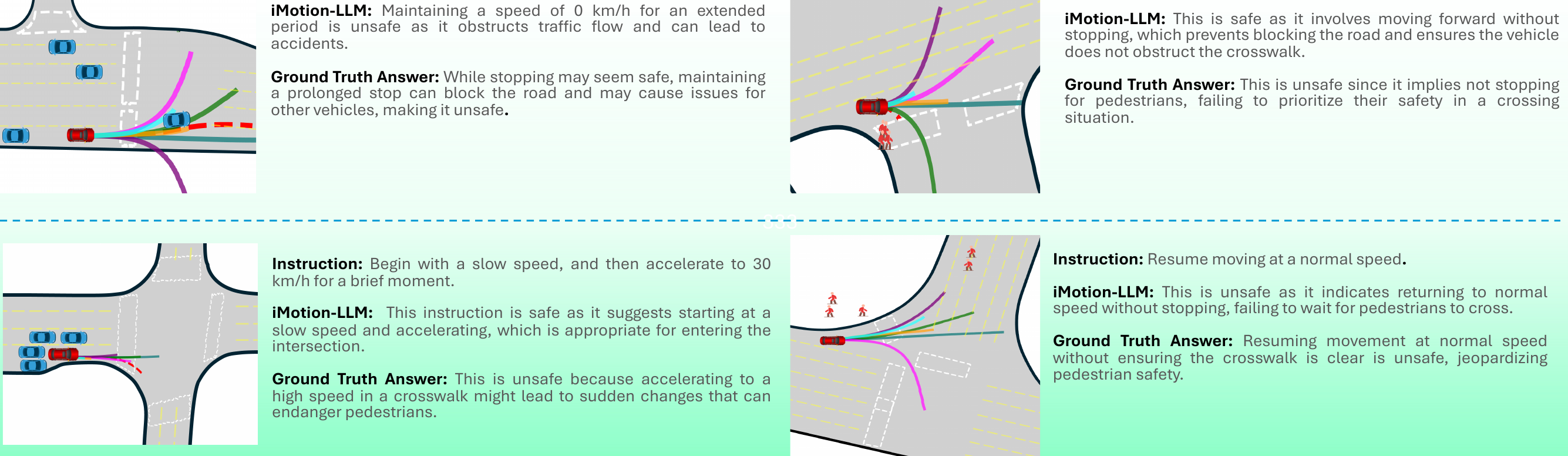}
\caption{Additional Qualitative Results.}
\label{fig_additiona_qu}
\end{figure*}

\section{Time and Memory Analysis}
In this section, we present a latency and memory analysis of Conditional-GameFormer, GameFormer, \textit{iMotion-LLM}, and ProSim. All results were obtained using a single NVIDIA A100 GPU, except for ProSim, which we report as provided in its original paper. The analysis highlights the differences in computational requirements between pure trajectory prediction models and LLM-based approaches. The results are summarized in Table~\ref{tab:a100_latency_memory}.





\begin{table}[h!]
  \centering
  \caption{\textbf{Forward-Pass Latency and Memory on A100.} Comparison of inference speed, memory usage, and whether the model produces output text.}
  \label{tab:a100_latency_memory}
  \footnotesize
  \setlength{\tabcolsep}{5pt}
  \scalebox{0.8}{
    \begin{tabular}{lccc}
      \toprule
      \textbf{Model} & \textbf{Fwd Latency (ms)} & \textbf{Peak Mem (MB)} & \textbf{Output Text} \\
        \cmidrule(r){1-1}
        \cmidrule(r){2-2}
        \cmidrule(r){3-3}
        \cmidrule(r){4-4}
      iMotion-LLM (7B, w/ text)   & 2100 & 7697 & Yes \\
      iMotion-LLM (7B, no text)   & 250  & 7010 & No  \\
      iMotion-LLM (1B, w/ text)   & 1200 & 2890 & Yes \\
      iMotion-LLM (1B, no text)   & 130  & 2600 & No  \\
      GameFormer                  & 40   & 139  & No  \\
      C-GameFormer                & 37   & 139  & No  \\
      ProSim-Instruct             & 324  & --   & Yes \\
      \bottomrule
    \end{tabular}
  }
\end{table}

\section{Additional Results and Analysis}
\subsection{Cross-Dataset Generalization and Fine-tuning Strategies}
We experiment with different training strategies to assess the generalization of \textit{iMotion-LLM} when evaluated on Open-Vocabulary InstructNuPlan safe instructions with context. All models start from a pretrained Conditional-GameFormer on InstructWaymo, with the following fine-tuning setups:  
(a) fine-tuning on InstructWaymo only,  
(b) fine-tuning directly on Open-Vocabulary InstructNuPlan,  
(c) fine-tuning on a direction-based InstructNuPlan (prepared similarly to InstructWaymo), and  
(d) two-stage fine-tuning: first on direction-based InstructNuPlan, then on Open-Vocabulary InstructNuPlan.

Interestingly, while single-stage fine-tuning on Open-Vocabulary InstructNuPlan yields high IFR, it produces poor-quality trajectories with large minADE/minFDE. In contrast, fine-tuning on the direction-based InstructNuPlan results in more generalizable trajectory generation, even when the model is prompted at inference with free-form Open-Vocabulary instructions. Moreover, following this with a second stage of Open-Vocabulary fine-tuning further improves both trajectory quality and instruction following. Notably, fine-tuning solely on Direction InstructNuPlan produces better trajectory quality than training on Open-Vocabulary InstructNuPlan alone, but with much lower IFR, highlighting weaker generalizability.

\begin{table}[t!]
  \centering
  \caption{\textbf{Generalization Results on Open-Vocabulary InstructNuPlan Safe with Context Examples.} Comparison across Stage-1/Stage-2 setups, reporting minADE/minFDE and IFR.}
  \label{tab:instructnuplan_results_gen}
  \footnotesize
  \setlength{\tabcolsep}{6pt}
  \scalebox{0.8}{
    \begin{tabular}{l l c c}
      \toprule
      \textbf{Stage-1} & \textbf{Stage-2} & \textbf{minADE/FDE} ↓ & \textbf{IFR} $\uparrow$ \\
      \cmidrule(r){1-2}
        \cmidrule(r){3-3}
        \cmidrule(r){4-4}
      Direction InstructWaymo   & None                      & 4.12 / 7.82 & 52.92 \\
      Open-Vocab InstructNuPlan & None                      & 6.33 / 9.19 & 70.22 \\
      Direction InstructNuPlan  & None                      & 1.22 / 2.79 & 61.91 \\
      Direction InstructNuPlan  & Open-Vocab InstructNuPlan & 1.03 / 2.08 & 67.16 \\
      \bottomrule
    \end{tabular}
  }
\end{table}


\subsection{Experimental Comparison with Language Conditioned Model.}
We directly compared our proposed \textit{iMotion-LLM} with ProSim-Instruct on the InstructWaymo benchmark (Table~2), at which we instruct the same agent under the same scenarios with the same instructions. The instructions are already a subset of what ProSim-Instruct was trained on, and the input data format matches the ProSim-Instruct implementation using their published model checkpoint and style of prompting. ProSim-Instruct achieves similar instruction-following rates (IFR) on ground-truth instructions (GT-IFR) but suffers from degraded trajectory quality demonstrated by the minADE/minFDE performance, whereas \textit{iMotion-LLM} maintains high-quality trajectories by leveraging the pretrained Conditional-GameFormer. Unlike ProSim-Instruct, which focuses only on standard prediction metrics mainly tailored for realism, our evaluation demonstrates controllability. Specifically, \textit{iMotion-LLM} achieves 52.24\% IFR on other feasible directions (F-IFR) despite not observing such instructions during training, compared to only 24.91\% for ProSim-Instruct. Moreover, the gap between other feasible and infeasible IFR highlights the persistent difficulty of generalizing to alternative controllable driving modes, underscoring the broader challenge of instruction-conditioned controllability in trajectory generation.

\section{Closed-Loop Small-Scale Evaluation of \textit{iMotion-LLM}}
We evaluate \papername{} in a closed-loop setting using the NuPlan planning simulator with non-reactive agents. Instructions are generated online by taking the ego vehicle’s current lane as the reference path and deriving directional commands over a 30-meter horizon.  

\begin{figure*}[t!]
\centering
\includegraphics[width=1.0\linewidth]{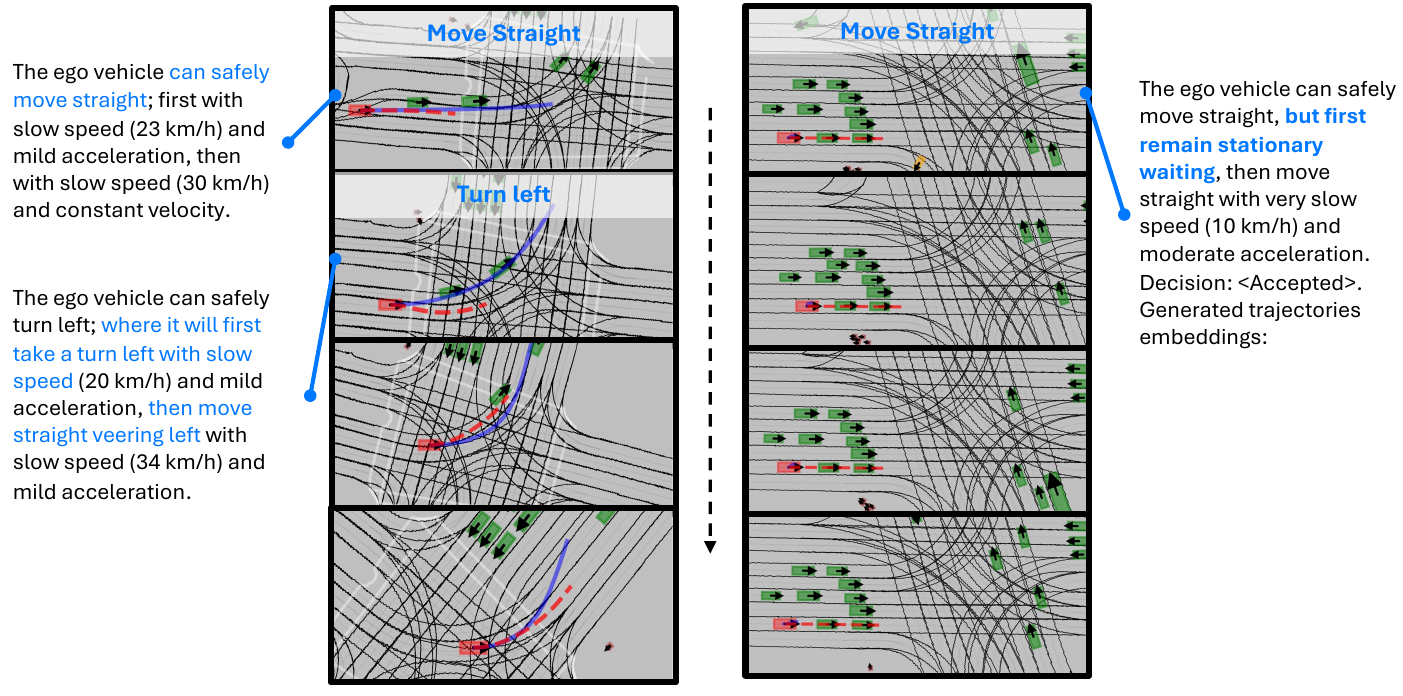}
\caption{
Two examples on the NuPlan closed-loop simulator. iMotion-LLM demonstrates generalizability to closed-loop settings, even under unsafe instructions, while providing textual justifications.
}
\vspace{-4mm}
\label{fig:closed_loop_exp}
\end{figure*}

Small-scale experiments (Figure~\ref{fig:closed_loop_exp}) demonstrate that \textit{iMotion-LLM} can operate effectively in this setup. In the left example, the model aligns well with the driving context, transitioning smoothly from going straight to turning left at an intersection while providing appropriate justifications. In the right example, where the instruction is vague and only safe if the vehicle waits first, the model both justifies and executes a stationary plan, demonstrating safety awareness. These results highlight the potential of LLM-based models to influence agent behavior in simulators, with future work extending to reactive agents and larger-scale evaluations.

\subsection{Text Justification Evaluation of iMotion-LLM on Open-Vocabulary InstructNuPlan}
While standard n-gram–based metrics such as BLEU, ROUGE, and METEOR are well established in natural language generation, they are known to be limited when applied to free-form reasoning text produced by large language models. Specifically, these metrics rely on surface-level n-gram overlap and therefore penalize valid paraphrases or alternative phrasings that capture the same semantics but use different wording. This makes them ill-suited for evaluating safety-critical reasoning, where conceptual fidelity and justification quality are more important than literal overlap.

To overcome these limitations, we adopted a rubric-guided evaluation using GPT-5 as a judge. This approach allows us to capture qualities that are not well reflected by string-matching metrics. We defined three essential aspects, each scored on a 0–10 scale. 
\begin{enumerate}
    \item \textbf{Safety and Risk Awareness:} whether the reasoning correctly identifies if the instruction leads to safe or unsafe driving and considers risks like collisions or loss of control.
    \item \textbf{Consistency with Instruction:} whether the reasoning faithfully follows the given instruction without contradiction or hallucination.
    \item \textbf{Clarity and Driving-Principle Justification:} whether the reasoning is concise, clear, and grounded in sound driving principles (e.g., control, speed, spacing).
\end{enumerate}

The rubric used for scoring was as follows: 
\begin{itemize}
    \item \textbf{0--2:} completely wrong or missing.
    \item \textbf{3--5:} partially correct with major gaps.
    \item \textbf{6--8:} mostly correct with minor issues or vague phrasing.
    \item \textbf{9--10:} fully correct, precise, and well-justified.
\end{itemize}

Using this framework, we evaluated four groups of data (safe/unsafe, with/without context). The results demonstrate both the robustness and high quality of the generated reasoning across settings as we show in Table ~\ref{tab:reasoning_quality_gpt}.

\begin{table}[h!]
  \centering
  \caption{\textbf{Instruction-Reasoning Quality Evaluation.} Average scores (0–10 scale) across safe/unsafe settings with and without context.}
  \label{tab:reasoning_quality_gpt}
  \footnotesize
  \setlength{\tabcolsep}{6pt}
  \begin{tabular}{l c}
    \toprule
    \textbf{Group} & \textbf{Average Score} \\
    \midrule
    Safe (no context)     & 7.80 \\
    Safe (with context)   & 7.98 \\
    Unsafe (no context)   & 7.62 \\
    Unsafe (with context) & 7.80 \\
    \midrule
    \textbf{Overall average} & \textbf{7.81} \\
    \bottomrule
  \end{tabular}
\end{table}


Breaking down by aspect, our model achieves 7.91 for Safety \& Risk Awareness, 8.11 for Consistency with Instruction, and 7.41 for Clarity \& Driving-Principle Justification. According to the rubric, these fall squarely in the "mostly correct with minor issues" band (6–8) and in many cases approach the "fully correct, precise, and well-justified" band (9–10). These results confirm that the generated reasoning is consistently safe, aligned with instructions, and well-grounded in driving principles.

\subsection{Single-Agent and Multi-Agent Predictions}
This work narrowly focuses on single-agent prediction, as agent-level trajectory controllability of causal trajectory can be viewed as different levels of complexity, with simpler instructions being necessary skill for a model capable of modeling more complex instructions. Simplest instructions can be in the form of high-level direction instructions for a single focal agent, despite that this is a single agent, this single agent can be any agent in the BEV map. The testing is independent of whether the vehicle is the actual ego vehicle during the recording of the data, or is another agent within the scene. This enables the model to generate marginal predictions of agents in a scene.

Multi-Agent joint prediction is common, similar to the original design of GameFormer and MTR, at which models perform well on the two interactive agents joint prediction on WOMD. Despite that, some studies questioned whether the joint modeling actually model semantics of complicated interactions between agents, at which they question both datasets and modeling [49].

In this work the focus was on single-agent, at which both simple high-level instructions were rigorously evaluated on InstructWaymo, and complex open-vocabulary high-level instructions were evaluated under different driving scenarios and conditions with feasibility and safety alignment evaluation.

In order to enable joint prediction, we used the original design of GameFormer for joint prediction, and similarly for the Conditional-GameFormer we consider generating a conditioning query per agent, using the same learnable embedding layer to be used in the decoding. To adapt \textit{iMotion-LLM}, we generate additional special tokens, to represent each target agent token in the Conditional-GameFormer decoder's input, as well two condition queries to be used in the decoding of trajectories, one for each agent. And the instruction is adapted such that it included instructing both agents (ego, and Agent-2), at which they are always the first two tokens in the scene tokens, and are pre identified as interactive agents by the WOMD.

We prepared experiments on joint prediction of two interactive agents, where we instruct both as we show in Table \ref{tab:joint_pred_results}. For the minADE and minFDE we show it for the ego agent only, and for the two agents, where for the IFR, we only show it for the ego agent. The results shows that Conditional-GameFormer actually get advantage of the joint prediction modeling, making the IFR almost 100\%, while improving the displacement errors over the non conditional model. Where our adaptation of iMotion-LLM does not show significant improvement over non conditional model, highlighting a modeling limitation that needs further exploration to take advantage of the pretrained Conditional-GameFormer performance on joint predictions. 

\begin{table}[t!]
  \centering
  \caption{\textbf{InstructWaymo Joint Prediction Results.} IFR and minADE/minFDE of the ego vehicle, or both interactive agents in joint prediction of two interactive agents.}
  \label{tab:joint_pred_results}
  \footnotesize
  \setlength{\tabcolsep}{5pt}
  \resizebox{\linewidth}{!}{
    \begin{tabular}{lcccc}
      \toprule
      \textbf{Model} & \textbf{minADE/FDE (Ego)} ↓ & \textbf{minADE/FDE (Both)} ↓ & \textbf{GT-IFR} ↑ & \textbf{F-IFR} ↑ \\
      \cmidrule(r){1-1}
      \cmidrule(r){2-2}
      \cmidrule(r){3-3}
      \cmidrule(r){4-4}
      \cmidrule(r){5-5}
      GameFormer   & 0.84/1.79 & 1.08/2.36 & 74.97 & 10.20 \\
      C-GameFormer & 0.72/1.33 & 0.94/1.91 & 96.70 & 52.20 \\
      iMotion-LLM  & 0.88/1.83 & 1.33/2.72 & 75.54 & 13.43 \\
      \bottomrule
    \end{tabular}
  }
\end{table}

\section{More Details on IFR Calculation}
\begin{figure}[h!]
\centering
\includegraphics[width=1.0\linewidth]{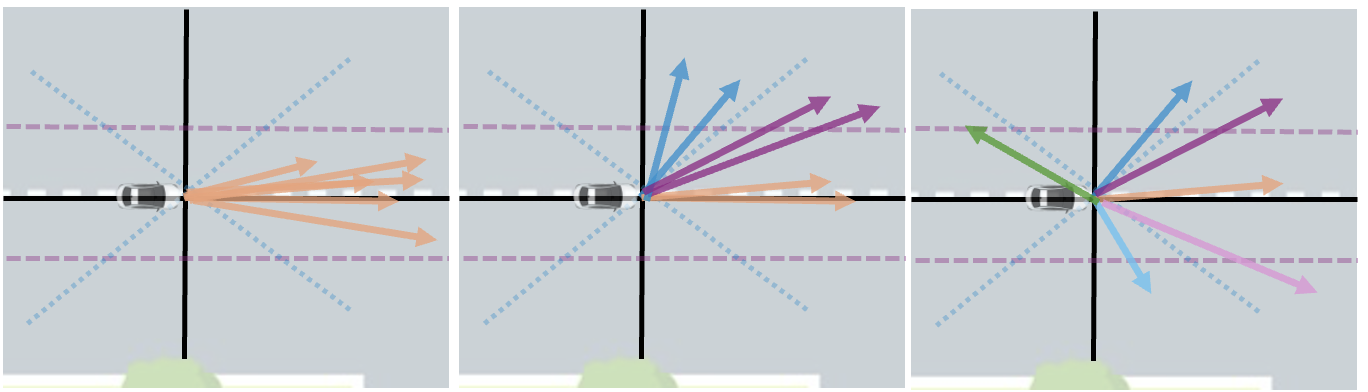}
\caption{
Illustrative examples of IFR calculation given an instruction of moving straight, with the first example scoring 100\%, the second 33.33\%, and the last 16.66\%.}
\vspace{-4mm}
\label{ifr_example_figure}
\end{figure}

To compute IFR, we consider the generation of six future trajectories for a given agent along with a reference directional instruction. For each example, we count how many of the generated trajectories follow the specified reference direction. These counts are then aggregated across all examples, where the accuracy for each direction is averaged separately over the corresponding samples. Finally, the macro average across all directions is reported as the overall IFR. Figure~\ref{ifr_example_figure} illustrates the IFR calculation for the instruction \textit{move straight}. In the left example, all six generated trajectories follow the instructed direction. In the middle, only two out of six trajectories comply, while in the rightmost example, just one out of six aligns with the instruction.

\section{More Details on Training}

We train both non-conditional and conditional baselines from scratch using 4×V100 GPUs. GameFormer and C-GameFormer are trained for 30 epochs, while MTR and C-MTR are trained for 15 epochs. For iMotion-LLM, we fine-tune starting from a pretrained C-GameFormer for 0.25 epochs (6,726 iterations) using 3×A100 GPUs, as longer fine-tuning did not yield further improvements. On the Open-Vocabulary InstructNuPlan dataset, iMotion-LLM is fine-tuned for 1 epoch (7,194 iterations). For training efficiency, we use 4-bit bfloat16 fine-tuning for iMotion-LLM. and we used AdamW optimizer with a learning rate of 1e-4, a maximum gradient norm of 10, and a cosine annealing scheduler with 0.1 epochs of warm-up. We apply a LoRA dropout of 0.05, batch size of 16, LoRA rank of 32, and LoRA alpha of 16. Additionally, we use a higher learning rate for the projection and mapping modules to accelerate adaptation.

\section{Conditional GameFormer and iMotion-LLM Training Pseudocodes}
This section presents the training pseudocode of Conditional GameFormer (Algorithm 1), adapted from GameFormer, with only lines 15 and 21 as the additional steps. Algorithm 2 shows the training pseudocode of \papernameAbbrev, which builds on top of Conditional GameFormer using a pretrained Conditional GameFormer checkpoint. Lines 21 and 22 in Algorithm 2 are highlighted to indicate the integration of Conditional GameFormer.

\label{appendix:pseudo}
\begin{algorithm}[!h]
\scriptsize
    \SetKwInOut{Input}{Input}
    \SetKwInOut{Output}{Output}

    \Input{
    $C_{instruction}\in \mathbb{Z}$: \text{Instruction category};
    $N_a$: \text{Num. agents}; 
    $d_a$: \text{Num. state features};
    $N_m$: \text{Num. map lanes};
    $N_p$: \text{Num. points per lane};
    $d_m$: \text{Num. map features};
    $d_{scene}$: \text{latent dimension}; 
    $t_{obs}=11$: \text{Observed time steps}; 
    $t_{pred}=80$: \text{To predict time steps}; 
    $t_{select}=[29, 49, 79]$: \text{Selected time steps}; 
    $N_{pred}$: \text{Two Agents to predict};
    $M$: \text{Num. modalities (futures)}; 
    $\mathbf{Agents} \in \mathbb{R}^{N_a\times t_{obs} \times d_a}$ : history states ; $\mathbf{Maps} \in \mathbb{R}^{N_{pred} \times N_m\times N_p \times d_m}$ ; 
    $N$: \text{Num. scene embeddings};
    }
    \vspace{0.5em}
    \Output{$\mathbf{Pred} \in \mathbb{R}^{M\times N_{pred}\times t_{pred}\times 4}$: prediction GMM parameters ($\mu_x, \mu_y, \sigma_x, \sigma_y$), where ($\mu_x, \mu_y$) are the 2D trajectory centers}
    \vspace{0.5em}
    $queried\_agents$ $\gets$ [0, 1, ..., $N_{pred}-1$]\tcp*{\scriptsize Target agents, [0,1] for two agents}
    $queried\_modalities$ $\gets$ $[0, 1, ..., M-1]$\tcp*{\scriptsize M modalities}
    \vspace{0.5em}
    $S \gets [$ $]$\tcp*{\scriptsize Initialize scene tokens empty list of embeddings}

    \For{each agent\_state in agents\_history}{
        agent\_emb $\gets$ Motion\_Encoder(agent\_state)\tcp*{\scriptsize Encode agent state}
        $S \gets S \cup \{agent\_emb\}$\tcp*{\scriptsize Append agent embedding to $S$}
    }

    \For{each map\_feature in map\_features}{
        map\_emb $\gets$ Map\_Encoder(map\_feature)\tcp*{\scriptsize Encode map feature}
        $S \gets S \cup \{map\_emb\}$\tcp*{\scriptsize Append map embedding to $S$}
    }

    $S \gets$ selfAttention($S$)\tcp*{\scriptsize Apply fusion self-attention encoder (Scene Encoder)}

    $K, V \gets S$\tcp*{\scriptsize Use $S$ as the keys and values of the trajectory decoder}
    \vspace{0.5em}

    $Q \gets [$ $]$\tcp*{\scriptsize Initialize $Q$}
    $q\_instruction$ $\gets$ Embedding($C_{instruction}$) \tcp*{\scriptsize \textbf{\textcolor{red}{Learnable instruction query (proposed)}}}
    
    \For{each agent\_number in queried\_agents}{
        q\_agent $\gets$ Embedding(agent\_number)\tcp*{\scriptsize agent query}
        \For{each modality\_number in queried\_modalities}{
            $q\_modality$ $\gets$ Embedding(modality\_number)\tcp*{\scriptsize Modality query}
            $q\_motion$ $\gets q\_agent + q\_modality$\tcp*{\scriptsize Combine queries}
            {$q\_motion \gets q\_motion + q\_instruction$}\tcp*{\scriptsize \textbf{\textcolor{red}{Add instruction query (proposed)}}}
            $Q \gets Q \cup \{q\_motion\}$\tcp*{\scriptsize Append motion query to $Q$}
        }
    }
    
    output\_features $\gets$ Multimodal\_Trajectory\_Decoder($Q, K, V$);

    \textbf{Pred}, Scores $\gets$ MLP(output\_features), MLP(output\_features)\tcp*{\scriptsize Get multimodal trajectories and modality scores}

    NLL\_loss $\gets$ NLL(Pred[best\_mode, :, $t_{select}$], ground\_truth\_2D)
    
    \textbf{gmm\_loss} $\gets$ NLL\_loss - CrossEntropy(Scores, best\_mode)

\caption{\scriptsize The pseudocode of Conditional-GameFormer (C-GameFormer).}

\end{algorithm}

\newpage
\begin{algorithm}[!h]
    \scriptsize
    \SetKwInOut{Input}{Input}
    \SetKwInOut{Output}{Output}

    \Input{
    Same inputs as C-GameFormer (Algorithm-1); 
    \\$T_{I}$: \text{Text input instruction};
    
    }
    \vspace{0.5em}
    \Output{Same output as C-GameFormer (Algorithm-1); \\Output Text}
    \vspace{0.5em}
    $queried\_agents$ $\gets$ [0, 1, ..., $N_{pred}-1$]\tcp*{\scriptsize Target agents, [0,1] for two agents}
    $queried\_modalities$ $\gets$ $[0, 1, ..., M-1]$\tcp*{\scriptsize M modalities}
    \vspace{0.5em}
    $S \gets $ Scene\_Encoder(agents\_history, map\_features) \tcp{\scriptsize (3-12) in Algorithm-1}
    
    $\Tilde{S} \gets [$ $]$
    
    \For{each $S_{embedding}$ in S}{
    $\Tilde{S} \gets \Tilde{S}$ $\cup$ LLM\_Projection($S_{embedding}$) \tcp*{\scriptsize Projections from $\mathbb{R}^{1xd_{scene}} \Rightarrow \mathbb{R}^{1\times d_{LLM}}$}
    }
    \textcolor{blue}{emb\_T$_\text{I}$ $\gets$ LLM\_Tokenizer(T$_\text{I}$) \tcp*{\scriptsize Embeddings of input text $\Rightarrow \mathbb{R}^{N_{tokens}\times d_{LLM}}$}}
    LLM\_Input\_emb $\gets$ [emb\_T$_\text{I}$; $\Tilde{S}$] \tcp*{\scriptsize concatenating text and scene embeddings}

    \If{\textcolor{blue}{Training}}{
    hidden\_states, tokens, LLM\_loss $\gets$ LLM(LLM\_Input\_emb) \tcp*{\scriptsize Autoregressive output last hidden states, corresponding tokens, and LLM cross-entropy loss}
    
    generation\_hidden\_states $\gets$ select\_generation\_states(hidden\_states) \tcp*{\scriptsize Selecting tokens that correspond to $[I],[S_1],[S_2],...[S_N]$}
    }
    \If{\textcolor{blue}{Inference}}{
    \While{[I] not detected}{
        next\_token $\gets$ LLM(LLM\_Input\_emb) \tcp*{\scriptsize Autoregressive next token generation until the first trajectory generation token [I] is found.}
        LLM\_Input\_emb $\gets$ LLM\_Input\_emb $\cup$ next\_token\_emb \tcp*{\scriptsize Include the next token to generate the following one}
    }
    hidden\_states $\gets$ Masked\_Generation\_LLM(LLM\_Input\_emb) \tcp*{\scriptsize Forcing the generation of all tokens $[I],[S_1],[S_2],...[S_N]$}
    }
    
    \textcolor{blue}{K,V $\gets$ Scene\_Mapper($[[S_1],[S_2],...[S_N]]$) \tcp*{\scriptsize Mapping each token independently, replaces (Line 13) in Algorithm-1}}
    
    \textcolor{blue}{$q\_instruction$ $\gets$ Instruct\_Mapper([I]) \tcp*{\scriptsize Mapping instruction token to $q_{instruct}$, replaces (15) in Algorithm-1}}
     
    $q_{motion} \gets $ Embedding($queried\_agents, queried\_modalities$) \tcp*{\scriptsize Combined agents-modalities queries, (16-20) in Algorithm-1}
    \vspace{0.5em}
    $Q \gets $ $q_{motion} + q_{instruction}$ \tcp*{\scriptsize Combine queries, (Line-22) in Algorithm-1}
    
    
    output\_features $\gets$ Multimodal\_Trajectory\_Decoder($Q, K, V$);

    \textbf{Pred}, Scores $\gets$ MLP(output\_features), MLP(output\_features)

    NLL\_loss $\gets$ NLL(Pred[best\_mode, :, $t_{select}$], ground\_truth\_2D)
    
    gmm\_loss $\gets$ NLL\_loss - CrossEntropy(Scores, best\_mode)
    
    \textbf{iMotion\_loss} = LLM\_loss + gmm\_loss

\caption{\scriptsize The pseudocode of iMotion-LLM.}
\end{algorithm}
